\documentclass{article} 
\usepackage{i2025_conference,times}

\usepackage{amsmath,amsfonts,bm}

\def\eqref#1{equation~\ref{#1}}

\def\1{\bm{1}}

\DeclareMathAlphabet{\mathsfit}{\encodingdefault}{\sfdefault}{m}{sl}
\SetMathAlphabet{\mathsfit}{bold}{\encodingdefault}{\sfdefault}{bx}{n}

\usepackage{hyperref}
\usepackage{url}

\usepackage{graphicx} 
\usepackage{url}
\usepackage{graphicx}
\usepackage{amsmath,amssymb} 
\usepackage{ragged2e}
\usepackage{colortbl}
\usepackage{cite}
\usepackage{color}
\usepackage{wrapfig}
\usepackage{hhline}
\usepackage{epsfig}
\usepackage{graphicx}
\usepackage{amsmath}
\usepackage{amssymb}
\usepackage{multirow}
\usepackage{booktabs}
\usepackage{algorithm}
\usepackage{algorithmic}
\usepackage{arydshln}
\usepackage{colortbl}
\usepackage{enumitem}
\usepackage{siunitx}
\usepackage{placeins}
\usepackage{bm}
\usepackage{makecell}
\usepackage{multirow}  
\usepackage{array}     
\usepackage{array}
\usepackage{caption}
\usepackage{amssymb}
\usepackage{pifont}
\usepackage{subcaption}  

\usepackage{listings}

\newcommand{\thickhline}{\Xhline{2\arrayrulewidth}}
\newcommand{\pub}[1]{{\color{gray}{\tiny{[{#1}]}}}}
\newcommand{\renderby}[1]{{\textcolor[HTML]{181842}{\tiny{({#1})}}}}
\newcommand{\textgr}[1]{\textcolor[HTML]{006400}{\textbf{#1}}}

\title{3DIS-flux: Simple and Efficient Multi-Instance Generation with DiT Rendering}

\author{Dewei Zhou$^1$, Ji Xie$^1$, Zongxin Yang$^2$, Yi Yang$^1$ $^*$ \\
$^1$RELER, CCAI, Zhejiang University $^2$DBMI, HMS, Harvard University\\
\texttt{\{zdw1999,sanaka87,yangyics\}@zju.edu.cn} \\
\texttt{\{Zongxin\_Yang\}@hms.harvard.edu} \\
\texttt{ \small $^*$ corresponding author}
\\
\small{\texttt{ project page: } \url{https://limuloo.github.io/3DIS/}}\\
}

\iclrfinalcopy 
\begin{document}

\maketitle

\vspace{-3ex}
\begin{figure}[h]
        \vspace{-5mm}
	\centering
        \includegraphics[width=0.95\linewidth]{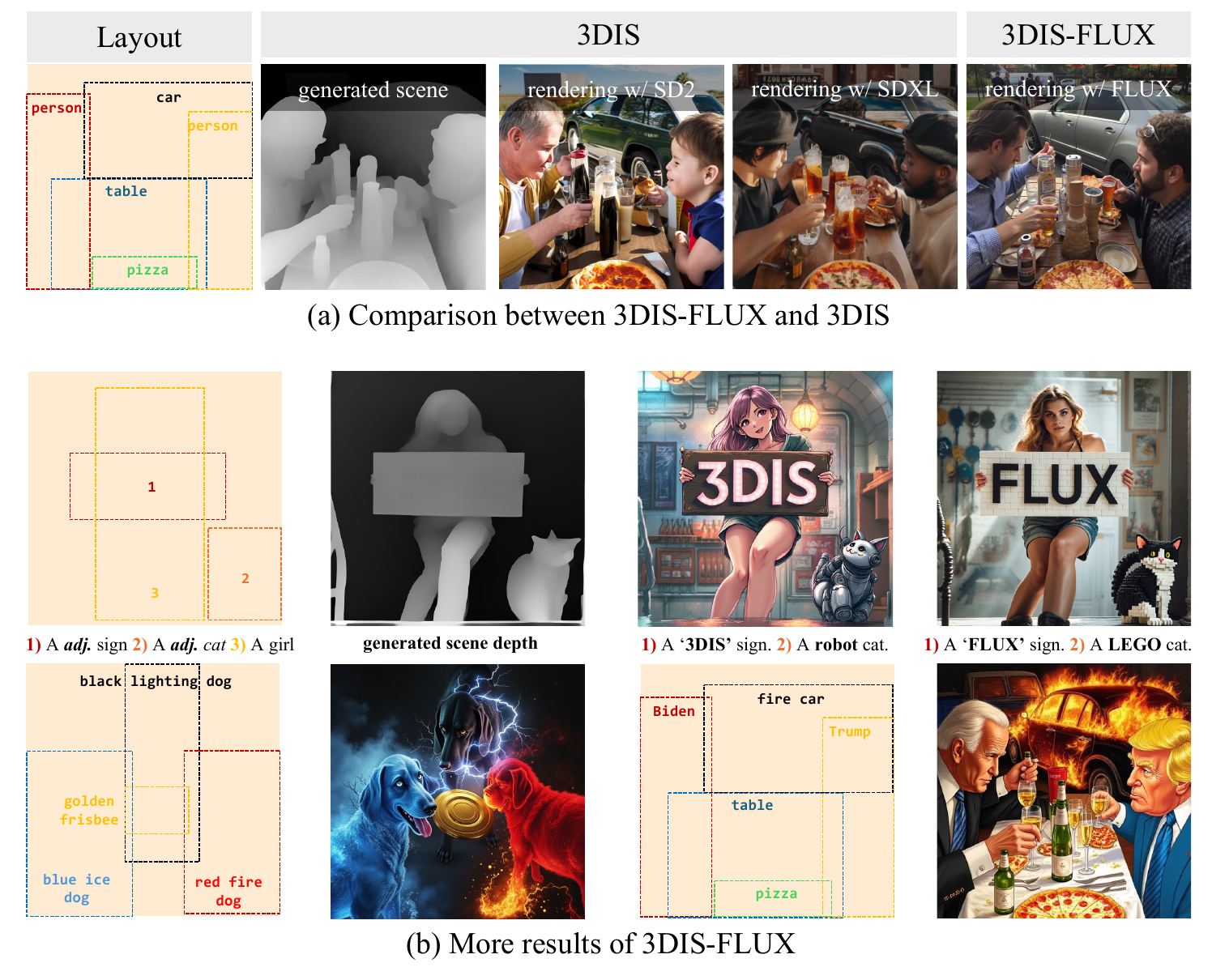}
	\caption{
    \textbf{Images generated using our 3DIS-FLUX.} Based on the user-provided layout, 3DIS~\citep{zhou20243dis} generates a scene depth map that precisely positions each instance and renders their fine-grained attributes without the need for additional training, using a variety of foundational models. Specifically, 3DIS-FLUX employs the state-of-the-art FLUX model for rendering, which is capable of producing superior image quality and offering enhanced control.
 }
 
\label{fig:teaser}
\end{figure}
\begin{abstract}

The growing demand for controllable outputs in text-to-image generation has driven significant advancements in multi-instance generation (MIG), enabling users to define both instance layouts and attributes. Currently, the state-of-the-art methods in MIG are primarily adapter-based. However, these methods necessitate retraining a new adapter each time a more advanced model is released, resulting in significant resource consumption. A methodology named Depth-Driven Decoupled Instance Synthesis (3DIS) has been introduced, which decouples MIG into two distinct phases: 1) depth-based scene construction and 2) detail rendering with widely pre-trained depth control models. The 3DIS method requires adapter training solely during the scene construction phase, while enabling various models to perform training-free detail rendering. Initially, 3DIS focused on rendering techniques utilizing U-Net architectures such as SD1.5, SD2, and SDXL, without exploring the potential of recent DiT-based models like FLUX. In this paper, we present 3DIS-FLUX, an extension of the 3DIS framework that integrates the FLUX model for enhanced rendering capabilities. Specifically, we employ the FLUX.1-Depth-dev model for depth map controlled image generation and introduce a detail renderer that manipulates the Attention Mask in FLUX's Joint Attention mechanism based on layout information. This approach allows for the precise rendering of fine-grained attributes of each instance. Our experimental results indicate that 3DIS-FLUX, leveraging the FLUX model, outperforms the original 3DIS method, which utilized SD2 and SDXL, and surpasses current state-of-the-art adapter-based methods in terms of both performance and image quality. Project Page: https://limuloo.github.io/3DIS/.
\end{abstract}

\section{Introduction}

With the rapid development of Diffusion Models~\citep{ddpm,ddim,pydiff,zhao2024learning,lu2024mace,lu2024vine,xie2024addsr}, contemporary models~\citep{stablediffusion,SD2,podell2023sdxl,flux} are capable of generating high-quality images. 
At the same time, there is a growing demand for more control over the generation process~\citep{controlnet,caphuman,lu2023tf,zhao2023wavelet}. One prominent area of research that has garnered increasing attention is Multi-Instance Generation (MIG)~\citep{migc,migc++,instancediffusion,gligen}, which seeks to ensure precise alignment of each instance’s position and attributes with user-defined specifications during the generation of multiple instances.

Current strategies for Multi-Instance Generation (MIG) can be broadly classified into three categories: 1) \textbf{Training-free methods}, such as Multi-Diffusion~\citep{multidiffusion} and RAG-Diffusion~\citep{ragdiff}, which employ multiple sampling steps for each instance and later merge them based on layout information to achieve spatial control. BoxDiff~\citep{boxdiff} and TFLCG~\citep{tflcg} define a score function to guide the model's sampling process, thereby enabling control over the layout. 2) \textbf{Adapter-based methods}, exemplified by MIGC~\citep{migc} and InstanceDiffusion~\citep{instancediffusion}, which introduce layout information by training additional attention layers atop pre-trained models. 3) \textbf{Text encoder fine-tuning methods}, such as Reco~\citep{reco} and Ranni~\citep{ranni}, which incorporate layout information directly into the input text and subsequently fine-tune the text encoder and the whole model to embed this spatial context into the generated output.

Adapter-based methods are currently widely used due to their ability to provide strong control without the need to train the entire model. However, these approaches face two main challenges: \textbf{1) the need to retrain on different models.} For example, methods like MIGC~\citep{migc}, GLIGEN~\citep{gligen}, and InstanceDiffusion~\citep{instancediffusion} were initially trained on SD1.5, but as more advanced models such as SDXL and SD3 emerged, techniques like IFAdapter~\citep{wu2024ifadapter} and CreatiLayout~\citep{zhang2024creatilayout} had to be retrained accordingly. This process is resource-intensive and can be particularly difficult for users with limited GPU access. \textbf{2) Large-scale instance-level annotations are often hard to obtain.} Each instance generation can be viewed as a Text-to-Image task, and high-quality instance-level data is generally more challenging to acquire than high-quality image-level data.

To address the aforementioned challenges, the Depth-Driven Decouple Instance Synthesis (3DIS)~\citep{zhou20243dis} approach introduces a novel framework for Multi-Instance Generation. Instead of directly generating RGB images, 3DIS first trains a layout-to-depth model to produce a scene depth map. Then, 3DIS utilizes widely pre-trained depth control models, which only require image-level annotations, to generate images based on the layout provided by the generated scene depth map. Finally, 3DIS employs a training-free method to precisely render the attributes of each instance. The 3DIS approach offers two key advantages: 1) It requires the training of only a depth generation model, which can ignore many fine-grained attributes during training and does not necessitate high-quality visual fidelity. 2) The training-free rendering in 3DIS enables the use of various pre-trained models and better preserves the generative capabilities of large pre-trained models.

The original 3DIS paper focused on the training-free rendering approach using models based on U-Net architectures, such as SD1.5~\citep{stablediffusion}, SD2~\citep{SD2}, and SDXL~\citep{podell2023sdxl}. With the advancement of diffusion model techniques, the Diffusion Transformer (DiT)~\citep{Li2024HunyuanDiTAP} architecture has demonstrated superior capabilities compared to traditional U-Net models. Specifically, FLUX has not only achieved significant improvements in image quality but has also enhanced control capabilities beyond those of previous models. As 3DIS is a flexible framework capable of quickly adapting to various new foundational models, we have extended it to propose 3DIS-FLUX, which leverages FLUX for training-free rendering, enabling stronger control and higher-quality image generation.
\begin{figure}[tb]
    \centering
    \includegraphics[width=1.0\linewidth]{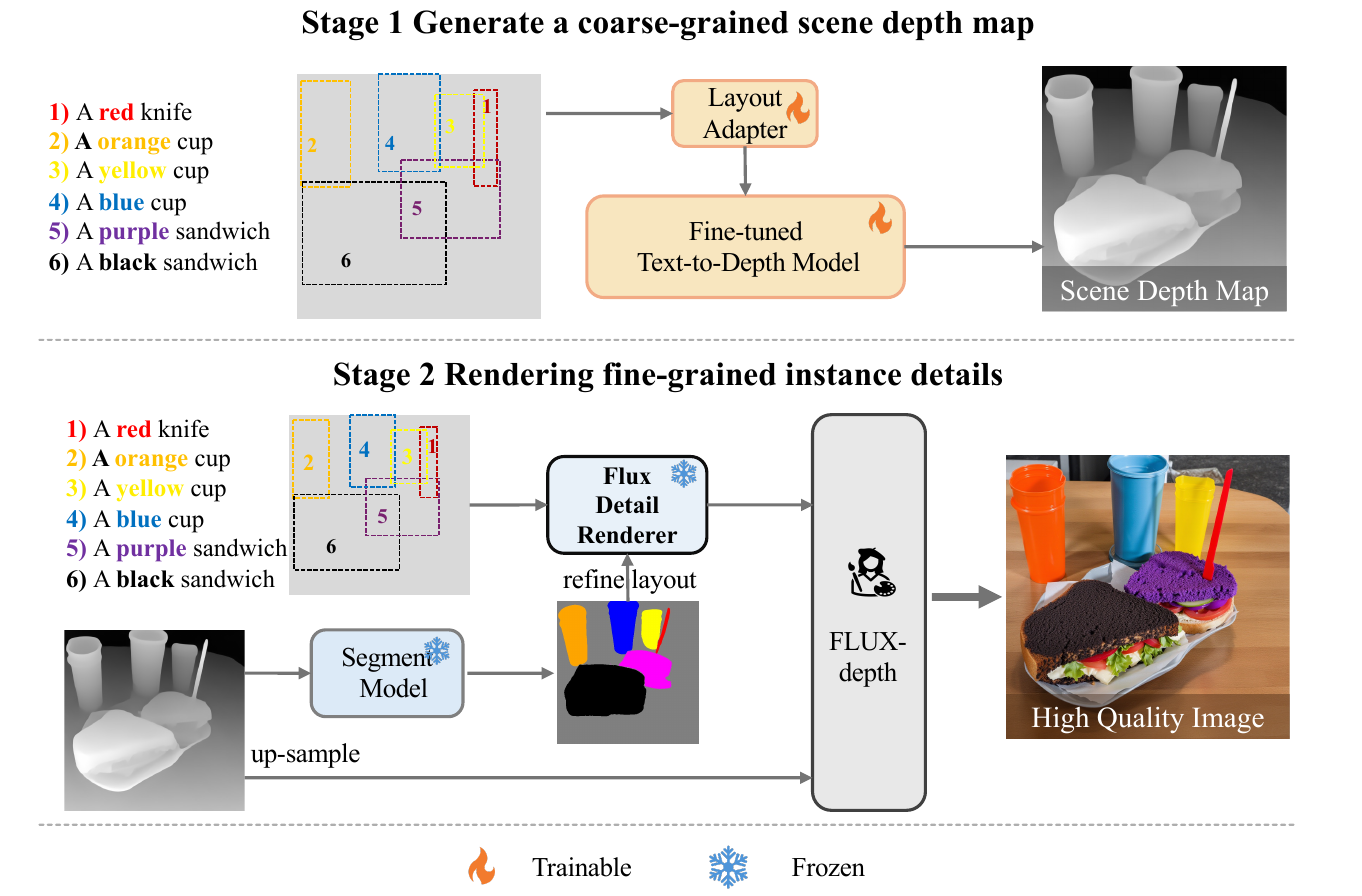}

\vspace{-1.5mm}
    \caption{\textbf{The overview of 3DIS-FLUX.} In line with 3DIS, the 3DIS-FLUX approach decouples image generation into two distinct stages: the creation of a scene depth map and the training-free rendering of high-quality RGB images using various generative models. 3DIS-FLUX utilizes the Layout-to-Depth model from 3DIS to generate the scene depth map, and subsequently employs the FLUX-depth model to render images based on the depth map. During this process, 3DIS-FLUX incorporates an Attention Controller to ensure the accurate fine-grained attributes of each instance.}
    \label{fig:overview}
    
\vspace{-4.5mm}

\end{figure}

The overview of 3DIS-FLUX is depicted in Fig.~\ref{fig:overview}. Consistent with the original 3DIS framework, 3DIS-FLUX initially generates a scene depth map using a layout-to-image model. Subsequently, we utilize the FLUX.1-depth-dev model for depth-to-image conversion. Through the application of the FLUX-depth model, alignment of the layout with the scene depth map is ensured. However, ensuring the accuracy of individual instance attributes still poses a challenge. To overcome this, we introduce a training-free Detail Renderer to achieve precise instance rendering in the Joint Attention~\citep{liu2024playground,dalva2024fluxspace,shin2024lianhuanhua} of the FLUX models. Specifically, we ensure that image tokens corresponding to each instance only attend to their respective text tokens. In the early steps, we mandate that each instance's image tokens only focus on their own image tokens. Moreover, since FLUX employs a T5 text encoder pre-trained exclusively on textual data—yielding embeddings devoid of visual information—we find that strict constraints on the attention map of text tokens within Joint Attention are crucial for successful rendering. Specifically, we restrict text tokens of each instance from attending to text tokens of other instances.

We conducted experiments on the COCO-MIG~\citep{migc} benchmark. The results show that by using the more powerful FLUX model for rendering, 3DIS-FLUX achieved a \textbf{6.9}\% improvement in Instance Success Ratio (ISR) compared to the previous 3DIS-SDXL. Compared to the training-free state-of-the-art (SOTA) method Multi-Diffusion, 3DIS-FLUX's improvement in ISR surpassed \textbf{41}\%. Against the SOTA adapter-based method, InstanceDiffusion, 3DIS-FLUX achieved a \textbf{12.4}\% higher ISR. Additionally, rendering with the FLUX model enabled our approach to demonstrate superior image quality compared to other methods.

\section{Method}
\label{method}

\subsection{Preliminaries}
FLUX~\citep{flux} is a recent state-of-the-art Diffusion Transformer (DiT) model that generates higher-quality images compared to previous models and demonstrates powerful text control capabilities. Given an input text, FLUX first encodes it into a text embedding using the T5 text encoder~\citep{T5}. This text embedding is then concatenated with the image embedding to perform joint attention. After several iterations of joint attention~\citep{liu2024playground,shin2024lianhuanhua,dalva2024fluxspace}, the FLUX model decodes the output image embedding to produce a high-quality image that corresponds to the input text.

\subsection{Problem Definition}

Multi-Instance Generation (MIG) requires the generation model to simultaneously produce multiple instances, ensuring that their positions and attributes align with the user's specifications. Given a layout \( P = \{p_1, p_2, \dots, p_n\} \) and the textual descriptions of the instances \( T = \{t_1, t_2, \dots, t_n\} \), MIG demands that each instance \( i \) be generated at the specified position \( p_i \), while visually matching the description \( t_i \). Additionally, the user provides a global text \( c \) that describes the entire scene, and the generated image must be consistent with this global text.

\subsection{Overview}
Fig.~\ref{fig:overview} illustrates the overview of 3DIS-FLUX. Similar to the original 3DIS, 3DIS-FLUX decouples Multi-Instance Generation into two stages: generating the scene depth map and rendering fine-grained details. In the first stage, 3DIS-FLUX employs the layout-to-depth model from 3DIS~\citep{zhou20243dis} to generate the corresponding scene depth map based on the user-provided layout. In the second stage, 3DIS-FLUX uses the FLUX.1-depth-dev~\citep{flux} model to generate an image from the scene depth map, thereby controlling the layout of the generated image. To further ensure that each instance is rendered with accurate fine-grained attributes, 3DIS-FLUX incorporates a detail renderer, which constrains the attention mask during joint attention between the image and text embeddings based on the layout information.

\subsection{FLUX Detail Renderer}

\noindent\textbf{Motivation.} Given a scene depth map generated in the first stage, FLUX.1-depth-dev model~\citep{flux} is capable of producing high-quality images that adhere to the specified layout. In scenarios involving only a single instance, users can achieve precise rendering by describing the instance through a single global image text. However, challenges arise when attempting to render multiple instances accurately with just one global text description. For example, in the case illustrated in Fig.~\ref{fig:overview}, rendering each ``cup" in the scene depth map with designated attributes using a description such as ``a photo of an orange cup, a yellow cup, and a blue cup" proves difficult. This approach frequently results in color inconsistencies, such as a cup intended to be blue being rendered as orange, with additional examples illustrated in Fig.~\ref{fig:ablation_detail_renderer}. Consequently, integrating spatial constraints into the joint attention process of the FLUX model is essential for the accurate rendering of multiple instances. To overcome these challenges, we introduce a simple yet effective FLUX detail renderer that significantly enhances the precision of such renderings.

\noindent\textbf{Preparation.} To render multiple instances simultaneously according to the user's descriptions, we encode not only the global image text \( c \) into \( f_c^{T5} \) but also the instance descriptions \(\{t_1, t_2, \ldots, t_n\}\) into \(\{f_1^{T5}, f_2^{T5}, \ldots, f_n^{T5}\}\). These encoded features are concatenated to form the final text embedding \( F = \text{concat}(f_c^{T5}, f_1^{T5}, \ldots, f_n^{T5})\), which is then input into the FLUX model's joint attention mechanism. Based on the user-provided layout \( P = \{p_1, p_2, \dots, p_n\} \), we determine the correspondence between image tokens and text tokens during the joint attention process. Since a scene depth map has already been generated in the first stage, we can opt to use the SAM~\citep{sam} model to further optimize the user's layout for more accurate rendering, as illustrated in Fig.~\ref{fig:overview}.

\noindent{\textbf{Controlling the Attention of Image Embedding.}} The FLUX model generates images through multi-step sampling. \textbf{1) The early steps} determine the primary attributes of each instance. Therefore, it is essential to strictly avoid attribute leakage by ensuring that the image token corresponding to instance \(i\) can only attend to the image tokens within the \(p_i\) region during joint attention and can only attend to its corresponding text token \(f_i^{T5}\). \textbf{2) In the later steps}, to ensure the quality of the generated image, we relax this constraint: each image token can attend to all other image tokens. Additionally, while attending to its corresponding text token \(f_i^{T5}\), it can also attend to the global text token \(f_c^{T5}\). We control these two phases by setting a threshold $\gamma$.

\noindent{\textbf{Controlling the Attention of Text Embedding.}} 
In the FLUX model, the T5 text encoder~\citep{T5}, which was pretrained solely on textual data, is employed to extract text encodings. This contrasts with previous methods that utilized the CLIP text encoder~\citep{CLIP}, which was pretrained using both text and image data. Notably, we observed that during the joint attention process, the T5 text embeddings inherently lack significant semantic information. If unconstrained, they are prone to inadvertently introducing incorrect semantic information. For instance, as demonstrated in Fig.~\ref{fig:ablation_T2T_control}, when the T5 text embeddings of ``black car" and ``green parking meter" are concatenated and input into FLUX's joint attention mechanism, allowing the ``green parking meter" tokens to attend to the ``black car" tokens results in the parking meter being predominantly black. Concurrently, it was found that FLUX was unable to successfully render the ``black car" at this stage. Therefore, it is imperative to impose constraints on the attention masks of the text tokens during joint attention to avoid such semantic discrepancies. We have discovered that imposing strict attention mask constraints on the text tokens of instances throughout all steps does not significantly affect the quality of the final generated image. Therefore, throughout all steps, we restrict the text token corresponding to \( f_i^{T5} \) during joint attention to only focus on the image tokens within the \( p_i \) area and to only attend to the text token of \( f_i^{T5} \) itself. For the text tokens of the global text token\( f_c^{T5} \), we do not apply significant constraints.

\section{Experiment}
\label{exp}

\subsection{Implement Details}

During the layout-to-depth phase, we employ the same method as used in the original 3DIS~\citep{zhou20243dis} approach. To incorporate depth control in image generation, we utilize the FLUX.1-depth-dev model~\citep{flux}. During the image generation process, we employ a sampling strategy of 20 steps. For images with a resolution of 512, the parameter $\gamma$ is set to 4. As the resolution increases, $\gamma$ is adjusted accordingly: it is set to 3 for images of 768 resolution and reduced to 2 for images of 1024 resolution.

\subsection{Experiment Setup}

\textbf{Baselines.} 
We compared our proposed 3DIS method with state-of-the-art Multi-Instance Generation approaches. The methods involved in the comparison include training-free methods: BoxDiffusion~\citep{boxdiff} and MultiDiffusion~\citep{multidiffusion}; and adapter-based methods: GLIGEN~\citep{gligen}, InstanceDiffusion~\citep{instancediffusion}, and MIGC~\citep{migc}.

\textbf{Evaluation Benchmarks.}
We conducted experiments on the COCO-MIG~\citep{migc} benchmark to assess a model's ability to control the position of instances and precisely render fine-grained attributes for each generated instance. For a comprehensive evaluation, each model generated 750 images in the benchmark.

\textbf{Evaluation Metrics.} We used the following metrics to evaluate the model: \textit{1) Mean Intersection over Union (MIoU)}, measuring the overlap between the generated instance positions and the target positions; \textit{2) Instance Success Ratio (ISR)}, calculating the proportion of instances that are correctly positioned and possess accurate attributes.

\begin{table*}[t!]
	\centering
	\caption{\textbf{Quantitative results on proposed COCO-MIG-BOX (\S\ref{sec:compare})}. $\mathcal L_{i}$ means that the count of instances needed to generate in the image is \textbf{i}.}\label{tab:coco_mig_box}
	\vspace{-3.5mm}
\centering
\setlength{\tabcolsep}{4.2pt}
\small
\renewcommand\arraystretch{1.1}
\begin{tabular}{rcccccccccccc}
\bottomrule[1pt]\rowcolor[HTML]{FAFAFA}
                                         & \multicolumn{6}{c}{Instance Success Ratio$\uparrow$}    & \multicolumn{6}{c}{Mean Intersection over Union$\uparrow$} \\

\cmidrule(lr){1-1} \cmidrule(lr){2-7} \cmidrule(lr){8-13}  
Method                                  & $\mathcal{L}2$         & $\mathcal{L}3$ & $\mathcal{L}4$ & $\mathcal{L}5$ & $\mathcal{L}6$ & $\mathcal{AVG}$ & $\mathcal{L}2$         & $\mathcal{L}3$ & $\mathcal{L}4$ & $\mathcal{L}5$ & $\mathcal{L}6$ & $\mathcal{AVG}$         \\ \toprule[0.8pt]

\rowcolor[HTML]{FDFFFD}  \multicolumn{13}{c}{\textit{Adapter rendering methods}} \\

\hline

\rowcolor[HTML]{FDFFFD} GLIGEN\hspace{0.30em}~\pub{CVPR23}\hspace{0.1em}            & 41.3 & 33.8                       & 31.8          & 27.0          & 29.5          & 31.3         & 33.7 & 27.6                       & 25.5          & 21.9          & 23.6          & 25.2         \\

\rowcolor[HTML]{FDFFFD} InstanceDiff\hspace{0.3em}~\pub{CVPR24}\hspace{0.1em} & 61.0 & 52.8                       & 52.4          & 45.2          & 48.7          & 50.5          & 53.8 & 45.8                       & 44.9          & 37.7          & 40.6          & 43.0  \\

\rowcolor[HTML]{FDFFFD}
MIGC\hspace{0.30em}~\pub{CVPR24}\hspace{0.1em}                                  & 74.8 & 66.2 & 67.4 & 65.3 & 66.1 & 67.1 & 63.0 & 54.7 & 55.3 & 52.4 & 53.2 & 54.7         \\

\hline
\toprule[0.8pt]
\rowcolor[HTML]{FCFEFF}  \multicolumn{13}{c}{\textit{\textbf{training-free} rendering}} \\

\hline

\rowcolor[HTML]{FCFEFF} TFLCG\hspace{0.0em}~\pub{WACV24}\hspace{0.1em}               & 17.2 & 13.5                       & 7.9          & 6.1          & 4.5          & 8.3          &  10.9 & 8.7                       & 5.1          & 3.9          & 2.8          & 5.3            \\
\rowcolor[HTML]{FCFEFF} BoxDiff\hspace{0.48em}~\pub{ICCV23}\hspace{0.1em}                                      & 28.4 & 21.4                       & 14.0          & 11.9          & 12.8          & 15.7          &  19.1 & 14.6 & 9.4 & 7.9 & 8.5 & 10.6          \\
\rowcolor[HTML]{FCFEFF} MultiDiff\hspace{0.40em}~\pub{ICML23}\hspace{0.1em}            &    30.6        & 25.3                       & 24.5          & 18.3          & 19.8          & 22.3          & 21.9        & 18.1                       & 17.3          & 12.9          & 13.9          & 15.8        \\
\hline
\rowcolor[HTML]{FCFEFF} 
{3DIS\hspace{1.3em}\renderby{SD1.5}}                             & 65.9&56.1&55.3&45.3&47.6 & 53.0 & 56.8&48.4&49.4&40.2&41.7  & 44.7         \\

\rowcolor[HTML]{FCFEFF}
{3DIS\hspace{1.3em}\renderby{SD2.1}}                             & 66.1&57.5&55.1&51.7&52.9 & 54.7 & 57.1&48.6&46.8&42.9&43.4  & 45.7         \\

\rowcolor[HTML]{FCFEFF}
{3DIS\hspace{1.3em}\renderby{SDXL}}                             & 66.1&59.3&56.2&51.7&54.1 & 56.0 & 57.0&50.0&47.8&43.1&44.6  & 47.0         \\

\rowcolor[HTML]{FCFEFF}
{3DIS-FLUX\hspace{1.3em}\renderby{FLUX}}                             & 76.4& 68.4 & 63.3 & 58.1 & 58.9 & 62.9 & 67.3 & 61.2 & 56.4 & 52.3 & 52.7  & 56.2         \\

\rowcolor[HTML]{FCFEFF} 
\multicolumn{1}{c}{vs. MultiDiff}                             & \textgr{+46} & \textgr{+43} & \textgr{+39} & \textgr{+40} & \textgr{+39} & \textgr{+41} & \textgr{+45} & \textgr{+43} & \textgr{+39} & \textgr{+39} & \textgr{+39} & \textgr{+40}

         \\

\hline
\toprule[0.8pt]
\rowcolor[HTML]{FFFDF5} 
\multicolumn{13}{c}{\textit{rendering w/ \textbf{off-the-shelf} adapters}} \\

\hline

\rowcolor[HTML]{FFFDF5} 
\multicolumn{1}{c}{3DIS+GLIGEN}                             & 49.4 & 39.7 & 34.5 & 29.6 & 29.9 & 34.1 & 43.0 &  33.8 & 29.2 & 24.6 & 24.5  & 28.8         \\

\rowcolor[HTML]{FFFDF5} 
\multicolumn{1}{c}{vs. GLIGEN}                             & \textgr{+8.1} & \textgr{+5.9} & \textgr{+2.7} & \textgr{+2.6} & \textgr{+0.4} & \textgr{+2.8} & \textgr{+9.3} & \textgr{+6.2} & \textgr{+3.7} & \textgr{+2.7} & \textgr{+0.9} & \textgr{+3.6}
         \\

\hline

\rowcolor[HTML]{FFFDF5}
\multicolumn{1}{c}{3DIS+MIGC}                             & 76.8 & 70.2 & 72.3 & 66.4 & 68.0 & 69.7 & 68.0 &  60.7 & 62.0 & 55.8 & 57.3  & 59.5         \\

\rowcolor[HTML]{FFFDF5} 
\multicolumn{1}{c}{vs. MIGC}                             & \textgr{+2.0} & \textgr{+4.0} & \textgr{+4.9} & \textgr{+1.1} & \textgr{+1.9} & \textgr{+2.6} & \textgr{+5.0} & \textgr{+6.0} & \textgr{+6.7} & \textgr{+3.4} & \textgr{+4.1} & \textgr{+4.8}        \\

\rowcolor[HTML]{FFF9E4} \toprule[0.8pt]

\end{tabular}
\vspace{-4.0mm}
\end{table*}

\begin{figure}[tb]
    \centering
    \includegraphics[width=1.0\linewidth]{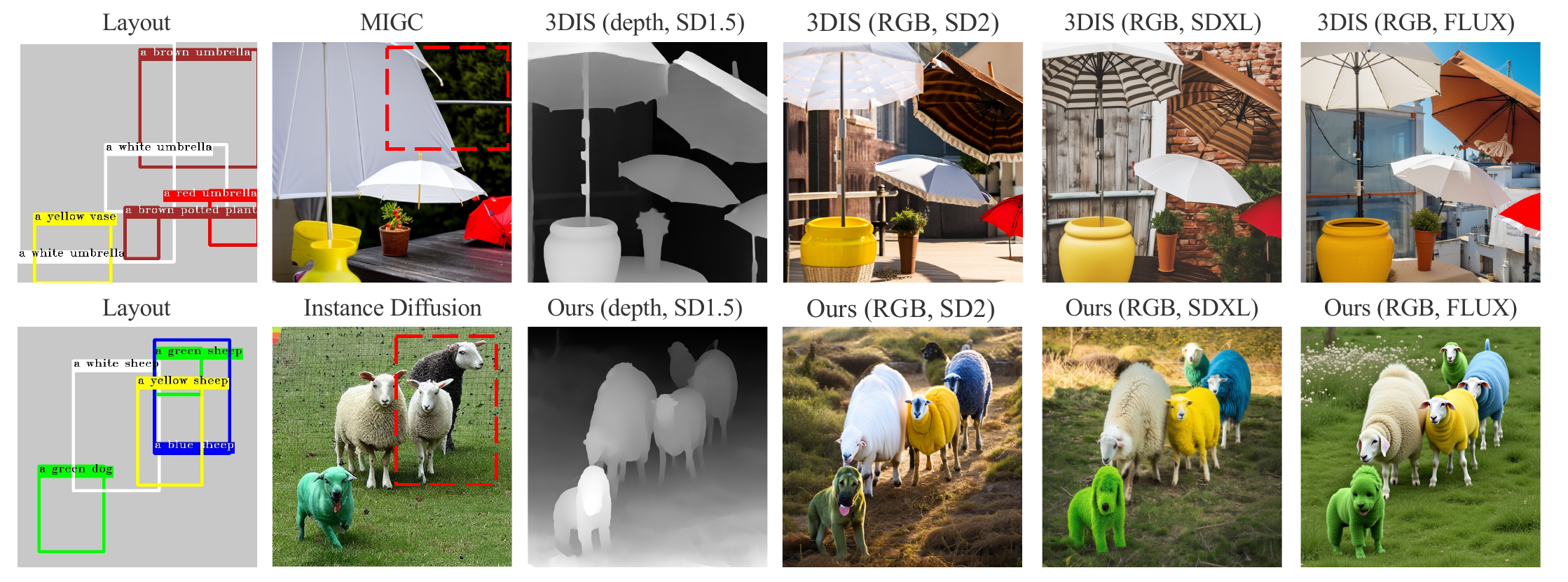}

\vspace{-3.5mm}
    \caption{\textbf{Qualitative results on the COCO-MIG (\S\ref{sec:compare})}. }
    \label{fig:coco_mig_vis}

\end{figure}
\begin{figure}[tb]
    \centering
    \includegraphics[width=0.98\linewidth]{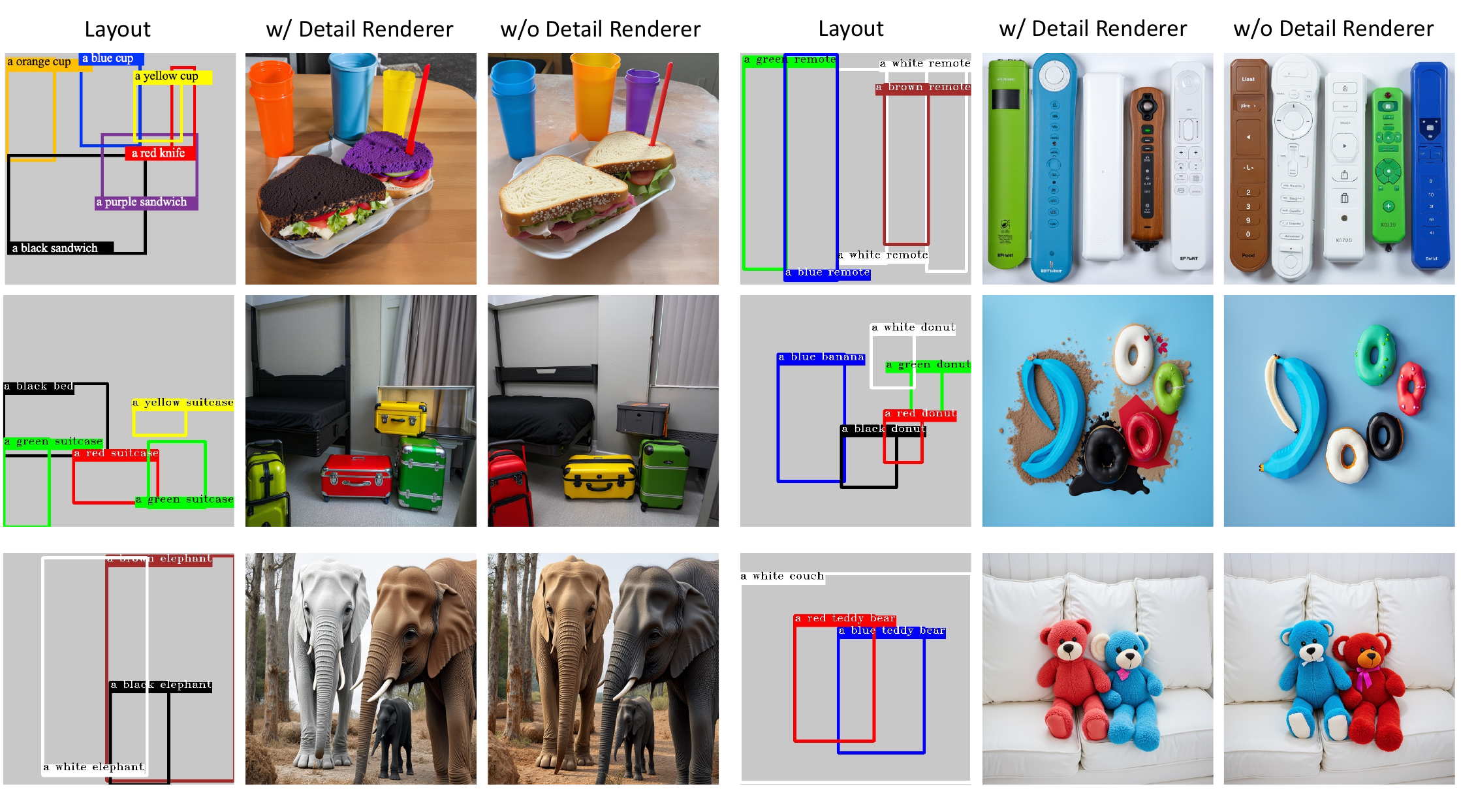}

\vspace{-3.5mm}
    \caption{\textbf{
Ablation Study on the FLUX Detail Renderer.}}
    \label{fig:ablation_detail_renderer}
    
\vspace{-3.5mm}

\end{figure}

\subsection{Comparison}
\label{sec:compare}

\textbf{Comparison with SOTA Methods.}
The results presented in Tab.~\ref{tab:coco_mig_box} demonstrate that the 3DIS method not only exhibits strong positional control capabilities but also robust detail-rendering capabilities.
Notably, the entire process of rendering instance attributes is \textbf{training-free} for 3DIS.
Compared to the previous state-of-the-art (SOTA) training-free method, MultiDiffusion, 3DIS-FLUX achieves a \textbf{41\%} improvement in the Instance Success Ratio (ISR).
Additionally, when compared with the SOTA adapter-based method, Instance Diffusion, which requires training for rendering, 3DIS-FLUX shows a \textbf{12.4\%} increase in ISR.
Importantly, the 3DIS approach is not mutually exclusive with existing adapter methods.
For instance, combinations like 3DIS+GLIGEN and 3DIS+MIGC outperform the use of adapter methods alone, delivering superior performance.
Fig.~\ref{fig:coco_mig_vis} offers a visual comparison between 3DIS and other SOTA methods, where it is evident that 3DIS not only excels in scene construction but also demonstrates strong capabilities in instance detail rendering.
Furthermore, 3DIS is compatible with a variety of base models, offering broader applicability compared to previous methods.

\textbf{Comparison of Rendering Across Different Models.} As shown in Tab.~\ref{tab:coco_mig_box}, employing a more robust model significantly enhances the success rate of rendering. For instance, rendering with the FLUX model achieves a \textbf{9.9}\% higher Instance Success Ratio compared to using the SD1.5 model.

\begin{figure}[tb]
    \centering
    \includegraphics[width=0.98\linewidth]{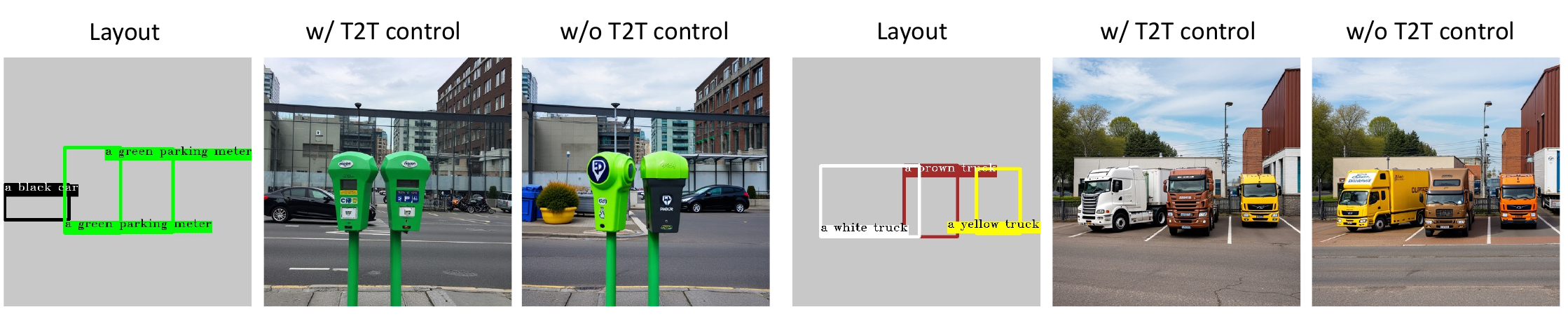}

\vspace{-3.5mm}
    \caption{\textbf{Ablation Study on Controlling Text-to-Text Attention in the FLUX Detail Renderer.
}}
    \label{fig:ablation_T2T_control}
    
\vspace{-3.5mm}

\end{figure}


\subsection{Ablation Study}
\label{sec:ablation}

\textbf{FLUX Detail Renderer.}
Results from Fig.~\ref{fig:ablation_detail_renderer} indicate that without employing a Detail Renderer to manage the Joint Attention process of the FLUX model, it becomes challenging to successfully render each instance in multi-instance scenarios. Additionally, data from Tab.~\ref{tab:ablation_render} demonstrates that the introduction of a Detail Renderer enhances the Instance Success Ratio (ISR) by \textbf{17.8}\% and the Success Ratio (SR) by \textbf{12.2}\%. Moreover, the results from Fig.~\ref{fig:ablation_detail_renderer} also suggest that incorporating a Detail Renderer does not significantly compromise image quality.

\setlength\intextsep{5pt}
\begin{wraptable}{r}{0.5\linewidth}
\centering
\caption{\textbf{Ablation study on FLUX detail renderer (\S\ref{sec:ablation}).}}
\label{tab:ablation_render}
\vspace{-6pt}
\small
\setlength\tabcolsep{4pt}
\renewcommand\arraystretch{1}
\begin{tabular}{|l||ccc|}
    \hline
    \thickhline
    \rowcolor[HTML]{FAFAFA}
 method  & ISR $\uparrow$ &  MIOU $\uparrow$ &  SR $\uparrow$ \\
    \hline\hline
    w/o I2I control & 55.4  & 49.9 & 21.1  \\ 
    w/o I2T control & 43.8  & 40.4 & 14.4 \\
    w/o T2I control & \textbf{63.4}  & \underline{56.2} & \underline{28.0}  \\
    w/o T2T control & 46.6  & 42.4 & 16.7 \\
    w/o detail renderer & 45.1  & 41.3 & 17.5  \\
    w/ all & \underline{62.9}  & \textbf{56.2} & \textbf{29.7} \\
    \hline
\end{tabular}
\vspace{-8pt}
\end{wraptable}

\noindent{\textbf{Controlling the Attention of Image Embedding.}} Results from Tab.~\ref{tab:ablation_render} show that in the Joint Attention mechanism, controlling each image token to focus solely on its corresponding instance description token (i.e., I2T control) is crucial for successfully rendering each instance, resulting in a \textbf{19.1}\% increase in the Instance Success Ratio (ISR). Additionally, restricting each instance's image tokens to only attend to other image tokens belonging to the same instance (i.e., I2I control) in the earlier steps of the process also leads to a significant improvement, enhancing the ISR by \textbf{7.5}\%.

\noindent{\textbf{Controlling the Attention of Text Embedding.}} In contrast to models such as SD1.5~\citep{stablediffusion}, SD2~\citep{SD2}, and SDXL~\citep{podell2023sdxl}, which utilize CLIP~\citep{CLIP} as their text encoder, FLUX employs a T5 text encoder~\citep{T5}. This encoder is exclusively pre-trained on textual data, resulting in embeddings that contain no visual information. Therefore, it becomes intuitively important in the joint attention mechanism to impose constraints on text tokens within multi-instance contexts. As demonstrated by the results in Tab.~\ref{tab:ablation_render} and Fig.~\ref{fig:ablation_T2T_control}, the absence of constraints on text tokens within joint attention mechanisms—permitting a text token from one instance to attend to text tokens from other instances—significantly undermines the rendering success rate, evidenced by a substantial decrease in ISR by \textbf{16.3}\%. Furthermore, our analysis reveals that adding constraints, where each instance's text tokens are restricted to only attend to their corresponding image tokens, does not result in a significant improvement.
\section{Conclusion}

We introduce 3DIS-FLUX, an extension of the prior 3DIS framework. The original 3DIS explored a training-free rendering approach using only the U-net architecture. In contrast, 3DIS-FLUX harnesses the state-of-the-art DiT model, FLUX, for rendering. Experiments conducted on the COCO-MIG dataset demonstrate that rendering with the more robust FLUX model allows 3DIS-FLUX to significantly outperform the previous 3DIS-SDXL method, and even surpass state-of-the-art Adapter-based MIG approaches. The success of 3DIS-FLUX underscores the flexibility of the 3DIS framework, which can rapidly adapt to a variety of newer, more powerful models. We envision that 3DIS will enable users to utilize a broader spectrum of foundational models for multi-instance generation and expand its applicability to more diverse applications.

\bibliography{i2025_conference}
\bibliographystyle{i2025_conference}

\newpage



\end{document}